\documentclass[10pt,twocolumn,letterpaper]{article}

\usepackage{iccv}
\usepackage{times}
\usepackage{epsfig}
\usepackage{graphicx}
\usepackage{amsmath}
\usepackage{amssymb}

\newcommand*{\affmark}[1][*]{\textsuperscript{#1}}

\makeatletter
\newcommand{\specificthanks}[1]{\@fnsymbol{#1}}
\makeatother

\usepackage[breaklinks=true,bookmarks=false]{hyperref}

\iccvfinalcopy 

\def\iccvPaperID{4055} 

\ificcvfinal\fi

\begin{document}

\title{Rethinking Coarse-to-Fine Approach in Single Image Deblurring}

\author{Sung-Jin Cho\thanks{equal contribution} \textsuperscript{ , 1} \qquad Seo-Won Ji\textsuperscript{\specificthanks{1}, 1} \qquad Jun-Pyo Hong\textsuperscript{1} \qquad Seung-Won Jung\thanks{corresponding author} \textsuperscript{ , 2} \qquad Sung-Jea Ko\textsuperscript{2}\\
	Department of Electrical Engineering, Korea University\\
	{\tt\small \affmark[1]\{sjcho, swji, jphong\}@dali.korea.ac.kr, \affmark[2]\{swjung83, sjko\}@korea.ac.kr}}

\maketitle
\ificcvfinal\thispagestyle{empty}\fi

\begin{abstract}
Coarse-to-fine strategies have been extensively used for the architecture design of single image deblurring networks. Conventional methods typically stack sub-networks with multi-scale input images and gradually improve sharpness of images from the bottom sub-network to the top sub-network, yielding inevitably high computational costs.
Toward a fast and accurate deblurring network design, we revisit the coarse-to-fine strategy and present a multi-input multi-output U-net~(MIMO-UNet).
The MIMO-UNet has three distinct features. 
First, the single encoder of the MIMO-UNet takes multi-scale input images to ease the difficulty of training. Second, the single decoder of the MIMO-UNet outputs multiple deblurred images with different scales to mimic multi-cascaded U-nets using a single U-shaped network. Last, asymmetric feature fusion is introduced to merge multi-scale features in an efficient manner. Extensive experiments on the GoPro and RealBlur datasets demonstrate that the proposed network outperforms the state-of-the-art methods in terms of both accuracy and computational complexity.
Source code is available for research purposes at \url{https://github.com/chosj95/MIMO-UNet}.
\end{abstract}

\section{Introduction}

Single image deblurring aims to recover a latent sharp image from a blurry image~\cite{fergus2006removing}. Even with the rapid development of camera modules in the last few decades, blur artifact still exists when camera and/or objects move. Blurry images are not only visually unpleasant but significantly degrade the performance of vision systems including surveillance~\cite{thorpe2013coprime} and autonomous driving systems~\cite{franke2000real}, necessitating accurate and efficient image deburring techniques.

\begin{figure}[]
	\includegraphics[width=\linewidth]{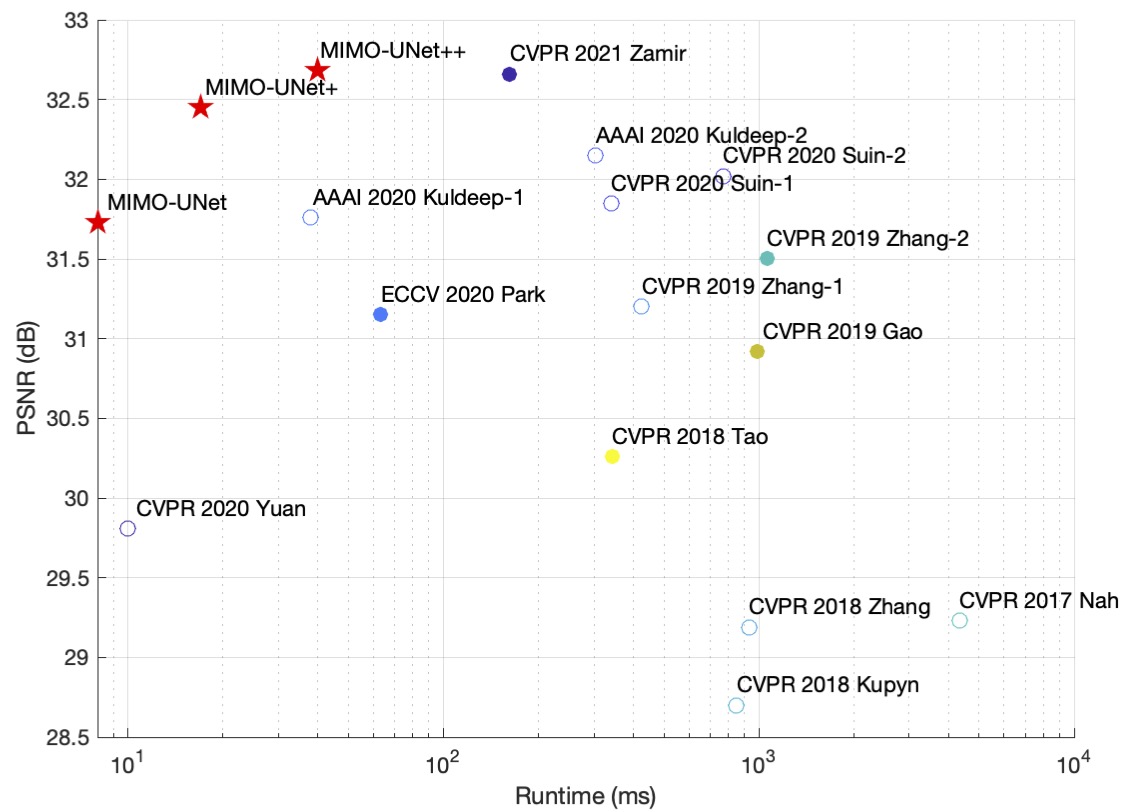}
	\caption{Comparison between the proposed and conventional methods in terms of the PSNR and runtime. The runtime of the methods is reported as the runtime measured using the released test code of each method on our environment (filled) and the runtime provided in each paper (blank).} 
	\label{fig: graph}
\end{figure}

\begin{figure*}[]
	\centering
	\includegraphics[width=0.99\linewidth]{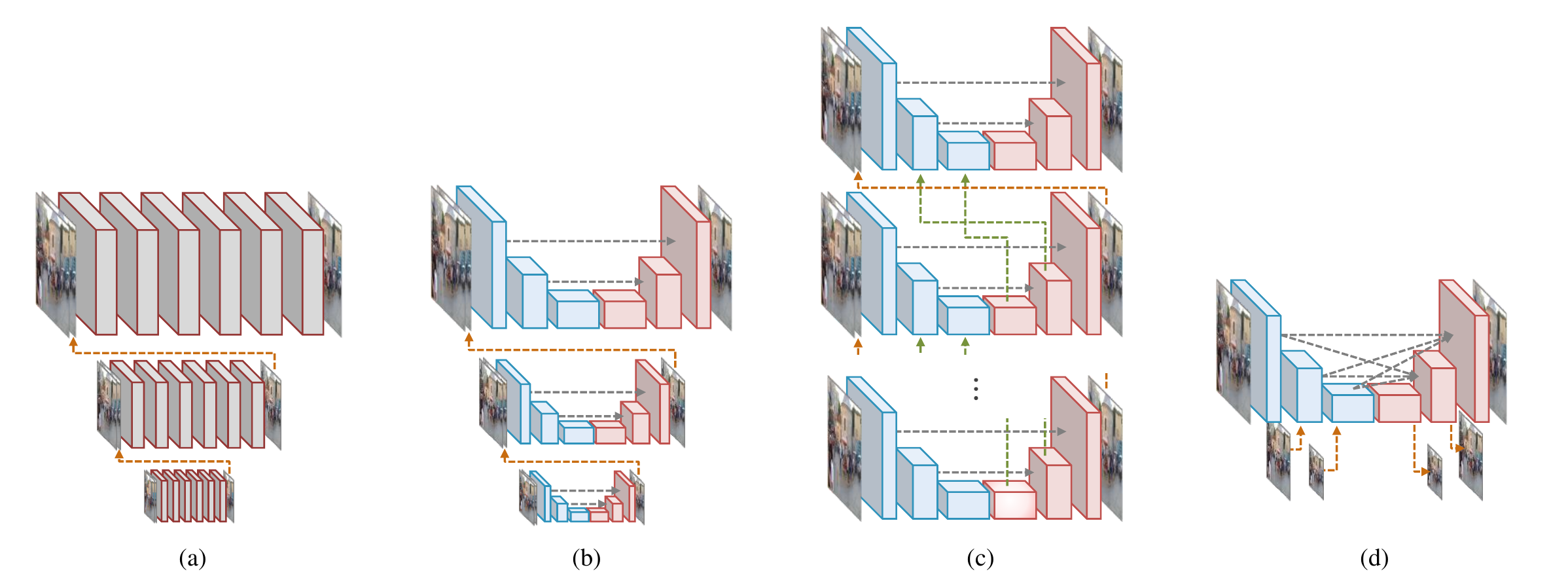}
	\caption{Comparison of coarse-to-fine image deblurring network architectures: (a)~DeepDeblur, (b)~PSS-NSC, (c)~MT-RNN, and (d)~proposed MIMO-UNet.}
	\label{fig: example}
\end{figure*}

Owing to the success of deep learning, convolutional neural network~(CNN)-based image deblurring methods have been extensively studied and showed promising performance.
Early CNN-based image deblurring methods~\cite{sun2015learning, hradivs2015convolutional, chakrabarti2016neural, schuler2015learning} commonly exploit CNN as a blur kernel estimator and construct two-stage image deblurring framework, \textit{i.e.}, CNN-based blur kernel estimation stage and kernel-based deconvoltion stage.
On the other hand, recent CNN-based image deblurring methods~\cite{nah2017, park2020multi, purohit2020region, tao2018scale, gao2019dynamic} aim to directly learn the complicated relationship between blurry-sharp image pairs in an end-to-end manner. As a pioneering technique, a deep multi-scale CNN for dynamic scene deblurring~(DeepDeblur)~\cite{nah2017} is introduced to directly regress a sharp image from a blurry image. DeepDeblur consists of multiple stacked sub-networks to handle multi-scale blur, where each sub-network takes a down-scaled image and gradually recovers a sharp image in a coarse-to-fine manner. Motivated by the success of DeepDeblur, various CNN-based image deblurring methods~\cite{park2020multi, purohit2020region, tao2018scale, gao2019dynamic} have been introduced with remarkable performance improvements.
Although these methods try to improve the deblurring performance in different aspects, their coarse-to-fine strategies are similar in that multiple sub-networks are stacked. In other words, a coarse-to-fine network design principle has proven to be effective in image deblurring. However, such efficiency comes at the cost of the inevitable increase in the computational complexity and memory usage, making the conventional methods difficult to be used for cost and time-sensitive environments such as mobile devices, vehicles, and robots. Recently, a light-weight CNN is presented for efficient single image deblurring~\cite{Yuan_2020_CVPR}. Specifically, by using optical flow and global motion of blurry images as extra supervision for network training, they design a shallower architecture compared to that of conventional deblurring networks. However, such shallow architecture failed in obtaining deblurring accuracy comparable to state-of-the-art methods.

In this paper, we revisit the coarse-to-fine scheme and present a novel deblurring network called multi-input multi-output UNet~(MIMO-UNet) that can handle multi-scale blur with low computational complexity. The proposed MIMO-UNet is a single encoder-decoder-based U-shaped network that has three distinct features. 

First, the single decoder of the MIMO-UNet outputs multiple deblurred images, and therefore we name our decoder as multi-output single decoder (MOSD). The MOSD is simple but can mimic conventional network architectures composed of stacked sub-networks and guide the decoder layers to gradually recover latent sharp images in a coarse-to-fine manner. Second, the single encoder of the MIMO-UNet takes multi-scale input images; thus, our encoder is called multi-input single encoder (MISE). Last, asymmetric feature fusion (AFF) is introduced to merge multi-scale features in an efficient manner. The AFF takes features from different scales and merges multi-scale information flow across the encoder and the decoder to improve the deblurring performance.
Extensive experiments demonstrate the superiority of the proposed MIMO-UNet compared to the state-of-the-art methods in terms of the PSNR as well as the computational complexity as shown in Figure~\ref{fig: graph}.

\section{Related works}

In this section, we review the conventional image deblurring methods that adopt a coarse-to-fine strategy.

\subsection{DeepDeblur}
As a pioneering work, DeepDeblur directly learns the relation between blurry-sharp image pairs in an end-to-end manner by adopting a coarse-to-fine strategy~\cite{nah2017}. Nah~\textit{et al.} also introduced the real-world image deblurring dataset named the GoPro dataset. Specifically, using a sequence of sharp images captured at 240 fps using a GoPro camera, a blurry image, $B$, is obtained by averaging successive sharp images as follow:
\begin{equation}
	B=\frac{1}{M} \sum_{i=0}^{M-1} S[i],
	\label{eq:blurgen}
\end{equation}
where $M$ and $S[i]$ represent the number of sampled sharp images and the $i^{th}$ sharp image, respectively. To construct a blurry and sharp image pair for training, the ground-truth sharp image for $B$ is chosen by selecting the middle image from the sampled sharp images.

To adopt a coarse-to-fine strategy in CNN for gradual recovery of latent sharp images, DeepDeblur uses multiple stacks of sub-networks as shown in Figure~\ref{fig: example}(a).
Each sub-network consists of a sequence of convolutional layers that maintains the spatial resolution of input feature maps. Different scales of input images are fed into the sub-networks, and the resultant image from a coarser scale sub-network is concatenated with the input of a finer scale sub-network to enable coarse-to-fine information transfer.
The reconstruction procedure of DeepDeblur is formulated as follows:
\begin{equation}
	{\hat{S}}_{n} = {\mathcal{H}}^\mathrm{D}_{\theta^\mathrm{D}_n}({B}_{n}; {(\hat{S}_{n+1})}^{\uparrow}) + B_{n},
	\label{Eq: convention1}
\end{equation}
where ${\mathcal{H}}^\mathrm{D}_{{\theta^\mathrm{D}_n}}$ is the $n^{th}$ sub-network of DeepDeblur parameterized by ${\theta^\mathrm{D}_n}$. $B_n$ and $\hat{S}_n$ are blurry and deblurred images at the $n^{th}$ scale, respectively, and $\uparrow$ denotes the up-sampling operation.

\begin{figure*}[!t]
	\centering
	\includegraphics[width=0.99\linewidth]{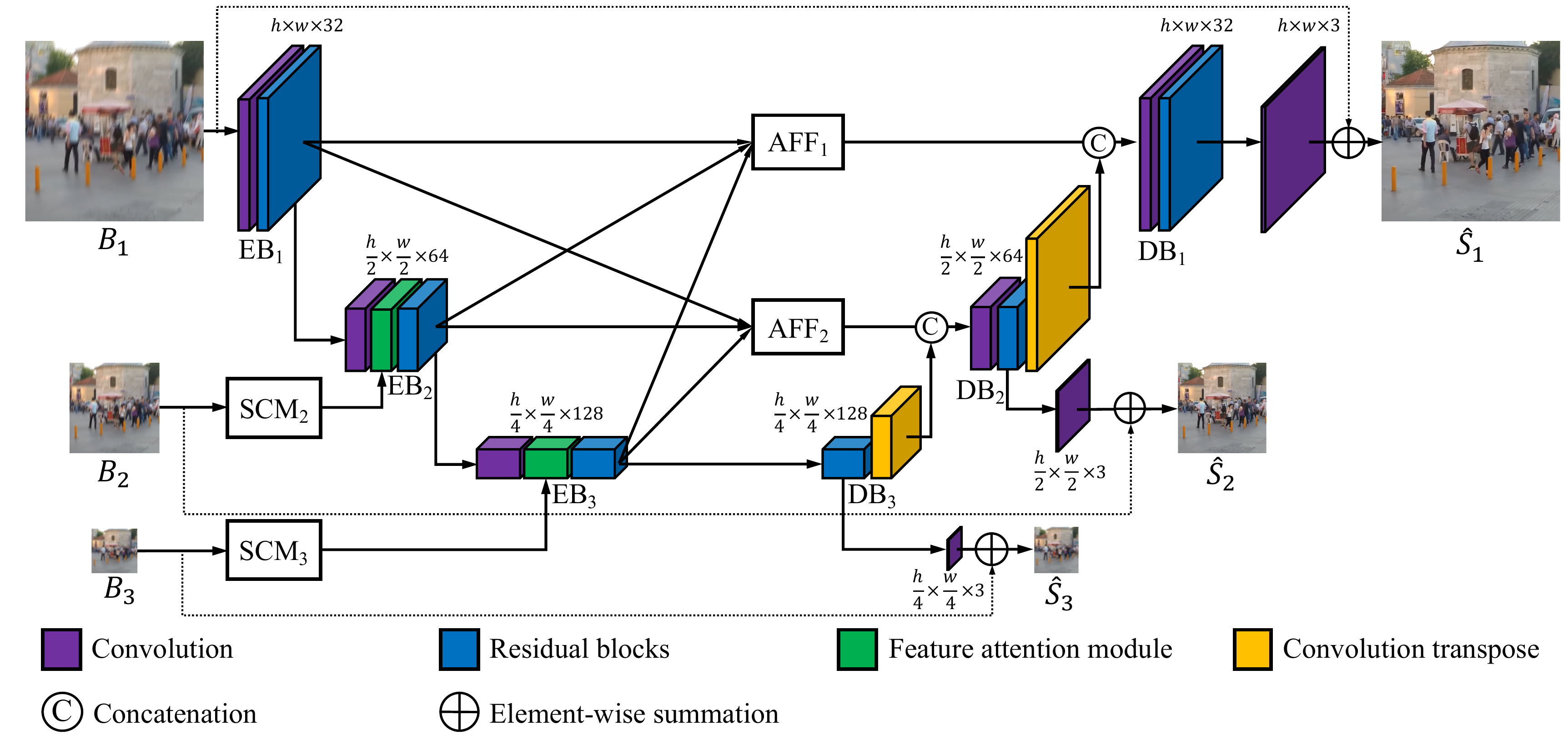}
	\caption{The architecture of the proposed network.}
	\label{fig: architecture}
\end{figure*}

\subsection{PSS-NSC}
Inspired by the success of DeepDeblur, Gao \textit{et al.} presented parameter selective sharing and nested skip connections~(PSS-NSC)~\cite{gao2019dynamic}. As shown in Figure~\ref{fig: example}(b), the architecture of PSS-NSC is similar to that of DeepDeblur, but has two distinct features.
First, each sub-network is structured as an encoder-decoder-based U-Net with symmetric skip connections that directly transfers the feature maps from the encoder to the decoder. Second, since every sub-network commonly aims to recover a sharp image from a blurry image, most network parameters are shared among sub-networks. Therefore, the memory requirement of PSS-NSC is significantly reduced, but the computational complexity is still demanding because the final sharp image is generated after passing through the three sub-networks. The reconstruction procedure of PSS-NSC is formulated as follows:
\begin{equation}
	{\hat{S}}_{n} = {\mathcal{H}}^\mathrm{P}_{({\theta}^\mathrm{P}_{n}, {\theta^\mathrm{P}})}({B}_{n}; {(\hat{S}_{n+1})}^{\uparrow}) + B_{n},
	\label{Eq: convention2}
\end{equation}
where ${\mathcal{H}}^\mathrm{P}_{({\theta}^\mathrm{P}_{n}, {\theta}^\mathrm{P})}$ represents the $n^{th}$ sub-network of PSS-NSC with exclusive parameter ${\theta}^\mathrm{P}_{n}$ and shared parameter ${\theta^\mathrm{P}}$.

\subsection{MT-RNN}
The network architecture of multi-temporal recurrent neural networks~(MT-RNN)~\cite{park2020multi} is illustrated in Figure~\ref{fig: example}(c). In MT-RNN, a single U-shaped network is repeated seven times, and the feature maps from the decoder at the previous iteration are transferred to the encoder at the next iteration as green colored arrows. For each iteration, MT-RNN is trained to predict an averaged image obtained using a different number of $M$ in Eq.~\ref{eq:blurgen}, where $M$ decreases as the iteration proceeds. 
Due to the repeated application of a single U-shaped network, MT-RNN has low memory usage but low runtime efficiency.
The reconstruction procedure of PSS-NSC is formulated as follows:
\begin{equation}
	\left\{ \hat{I}^{i}, F^{i}\right\}=\mathcal{H}^\mathrm{M}_{\theta^\mathrm{M}}\left(B^{i};\hat{I}^{i-1}, F^{i-1}\right),
\end{equation}
where $i$ refers to an iteration index. $\mathcal{H}^\mathrm{M}_{\theta^\mathrm{M}}$ is the network of MT-RNN parameterized by ${\theta^\mathrm{M}}$. $B^{i}$, $\hat{I}^{i}$, and $F^{i}$ are input blurry image, estimated latent image, and feature maps at the $i^{th}$ iteration, respectively.

\section{Proposed method}
We propose MIMO-UNet that fully exploits multi-scale features extracted from an input image. Figure~\ref{fig: architecture} shows the overall architecture of MIMO-UNet.
The architecture of MIMO-UNet is based on a single U-Net~\cite{ronneberger2015u} with significant modifications for efficient multi-scale deblurring. The encoder and decoder of MIMO-UNet are composed of three encoder blocks~(EBs) and decoder blocks~(DBs). The following subsections detail the three special features of MIMO-UNet, \textit{i.e.}, MISE, MOSD, and AFF.

\subsection{Multi-input single encoder}
It has been demonstrated that different levels of blur in images can be better handled from multi-scale images~\cite{michaeli2014blind, Liu2016CSVT}. Various CNN-based deblurring methods have also adopted this idea by taking a blurry image with a different scale as an input of each sub-network~\cite{nah2017, suin2020spatially, tao2018scale, gao2019dynamic}.

In our MIMO-UNet, not a sub-network but an EB takes a blurry image with a different scale as an input. In other words, in addition to the downsized feature extracted from the above EB, we extract the feature from the downsampled blurry image and then combine both features. By taking advantage of the complementary information from the downsized feature and the feature obtainable from the downsampled image, our EB is expected to handle diverse image blurs effectively. 
The use of multi-scale images as an input for a single U-Net has also proven to be effective in other tasks such as depth map super-resolution~\cite{guo2018hierarchical} and object detection~\cite{pang2019efficient}.

We first extract the features from the downsampled image using a shallow convolutional module~(SCM) as shown in Figure~\ref{fig: submodules}(a).
Considering efficiency, we use two stacks of $3 \times 3$ and $1 \times 1$ convolutional layers. We concatenate the features from the last $1 \times 1$ layer with the input $B_k$, and further refine the concatenated features using an additional $1 \times 1$ convolutional layer. The output of the SCM at the $k^{th}$ level is denoted as ${\rm{SCM}}_k^{{\rm{out}}}$, where we use SCM for the second and third levels as shown in Figure~\ref{fig: architecture}.

For the fusion of ${\rm{SCM}}_k^{{\rm{out}}}$ with the output of the $k-1^{th}$ level EB, ${\rm{EB}^{out}_{\it{k-1}}}$, we apply a convolutional layer with a stride of 2  to ${\rm{EB}^{out}_{\it{k-1}}}$, resulting in ${\left( {{\rm{EB}}_{k-1}^{{\rm{out}}}} \right)^ \downarrow }$. The two features ${\left( {{\rm{EB}}_{k-1}^{{\rm{out}}}} \right)^ \downarrow }$ and ${\rm{SCM}}_k^{{\rm{out}}}$ have the same size and thus can be fused. 
Here, we exploit a feature attention module~(FAM) to actively emphasize or suppress the features from the previous scale and learn the spatial/channel importance of the features from SCM.
We experimentally demonstrate that this module increases the performance compared to general feature fusion approaches as detailed in Sec.~\ref{subsec: abl}.

In particular, ${\left( {{\rm{EB}}_{k-1}^{{\rm{out}}}} \right)^ \downarrow }$ and ${\rm{SCM}}_k^{{\rm{out}}}$ are element-wise multiplied with each other, and then the multiplied features are passed through a $3 \times 3$ convolutional layer. The output of the $3 \times 3$ convolutional layer is expected to include complementary information for deblurring, and finally added to ${\left( {{\rm{EB}}_{k-1}^{{\rm{out}}}} \right)^ \downarrow }$ to be further refined through following residual blocks, where we used eight modified residual blocks~\cite{tao2018scale}.

\subsection{Multi-output single decoder}
In MIMO-UNet, different DBs have feature maps with different sizes. We consider that these multi-scale feature maps can be used to mimic multi-stacked sub-networks. Unlike the intermediate supervision at the sub-network as the conventional coarse-to-fine networks, we apply the intermediate supervision to each DB. The image reconstruction in each level can be formulated as follows:
\begin{equation}
	{\hat{S}}_{n}=\left\{ \begin{array}{l}
		o(\mathrm{{DB}}_n(\mathrm{{AFF}_{\it{n}}^{out}}; { \mathrm{{DB}_{\it{n}+1}^{out}}})) + B_{n}, ~ n = 1,2,\\
		o(\mathrm{{DB}}_n({ \mathrm{{EB}_{\it{n}}^{out}}})) + B_{n}, ~~~~~~~~~~~~~~~~~~ n = 3,
	\end{array} \right.
	\label{Eq: ours}
\end{equation}
where $\mathrm{{AFF}}^\mathrm{out}_n$, $\mathrm{{EB}}^\mathrm{out}_n$, and $\mathrm{{DB}}^\mathrm{out}_n$ are the outputs of the $n^{th}$ level asymmetric feature fusion~(AFF) module, EB, and DB, respectively.
Since the output of DB is a feature map not an image, mapping function $o$ is required for generating an intermediate output image, where we use a single convolutional layer.

\subsection{Asymmetric feature fusion}
\begin{figure}[t]
	\centering
	\includegraphics[width=0.95\linewidth]{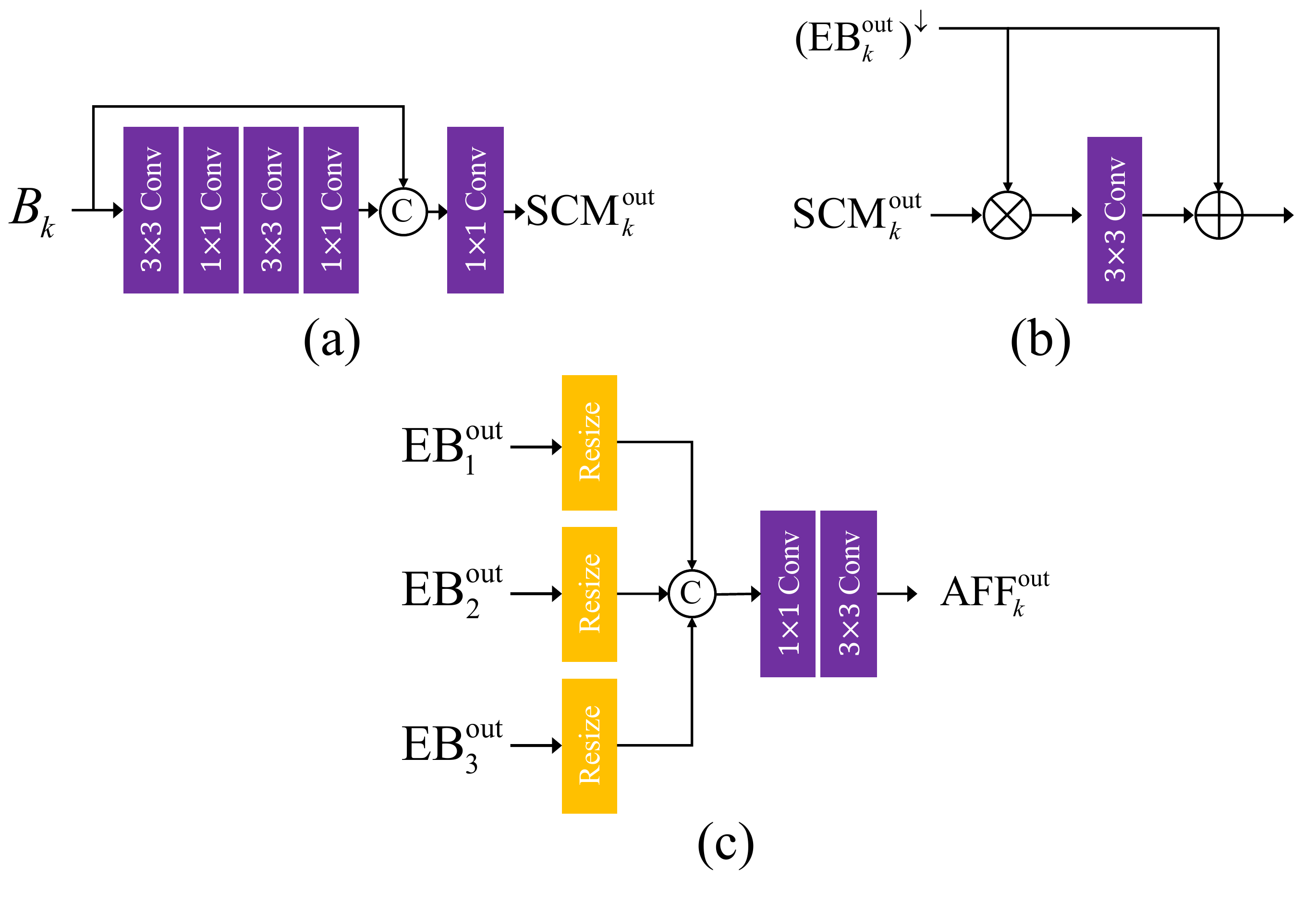}
	\caption{The structures of sub-modules: (a)~SCM, (b)~feature attention, and (c)~AFF.}
	\label{fig: submodules}
\end{figure}

\begin{figure*}[!t]
	\centering
	\includegraphics[width=\linewidth]{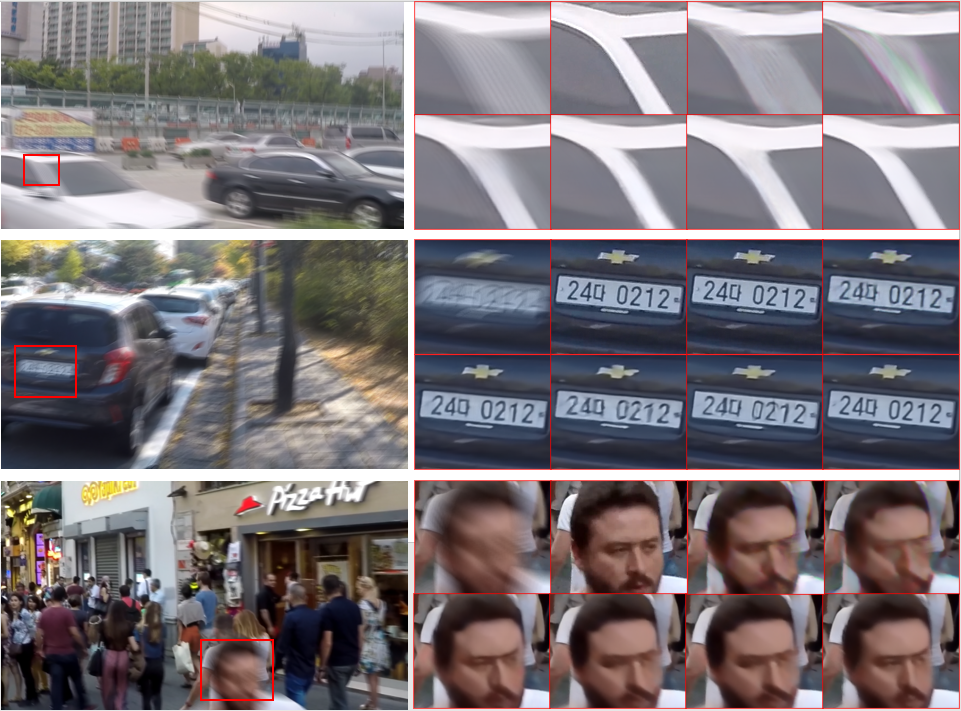}
	\caption{Several examples on the GoPro test dataset. For clarity, the magnified parts of the resultant images are displayed. From left-top to right-bottom: Blurry images, ground-truth images, and the resultant images obtained by SRN, PSS-NSC, DMPHN, MT-RNN, MPRNet, and MIMO-UNet++, respectively.}
	\label{fig: resultimage}
\end{figure*}

In most conventional coarse-to-fine image deblurring networks, only the features from the coarser-scale sub-network are used for the finer-scale sub-networks, making information flow inflexible. One exceptional method is to cascade the whole network in horizontal or vertical direction, allowing top-to-bottom and bottom-to-top information flow~\cite{zhang2019deep}.

Inspired by dense connection between intra-scale features~\cite{kim2018parallel}, we present an asymmetric feature fusion (AFF) module as shown in Figure~\ref{fig: submodules}(c) to allow information flow from different scales within a single U-Net. Each AFF takes the outputs of all EBs as an input and combines multi-scale features using convolutional layers. The output of the AFF is delivered to its corresponding DB.
More specifically, the first-level and second-level AFFs, ${\rm{AFF}}_{\rm{1}}$ and ${\rm{AFF}}_{\rm{2}}$, are formulated as follows:
\begin{equation}
	\begin{array}{l}
		{\rm{AFF}}_{\rm{1}}^{{\rm{out}}}{\rm{ = AF}}{{\rm{F}}_1}\left( {{\rm{EB}}_{\rm{1}}^{{\rm{out}}}{\rm{,}}{{\left( {{\rm{EB}}_{\rm{2}}^{{\rm{out}}}} \right)}^ \uparrow }{\rm{,}}{{\left( {{\rm{EB}}_{\rm{3}}^{{\rm{out}}}} \right)}^ \uparrow }} \right)\\
		{\rm{AFF}}_{\rm{2}}^{{\rm{out}}}{\rm{ = AF}}{{\rm{F}}_2}\left( {{{\left( {{\rm{EB}}_{\rm{1}}^{{\rm{out}}}} \right)}^ \downarrow }{\rm{,EB}}_{\rm{2}}^{{\rm{out}}}{\rm{,}}{{\left( {{\rm{EB}}_{\rm{3}}^{{\rm{out}}}} \right)}^ \uparrow }} \right)
	\end{array},
\end{equation}
where $\mathrm{{AFF}^{out}_{n}}$ represents the outputs of the $n^{th}$ AFF.
Up-sampling ($\uparrow$) and down-sampling ($\downarrow$) are applied such that the features from different scales can be concatenated. 
Each DB of MIMO-UNet can thus exploit multi-scale features, resulting the improved deblurring performance.

\begin{figure*}[!t]
	\centering
	\includegraphics[width=\linewidth]{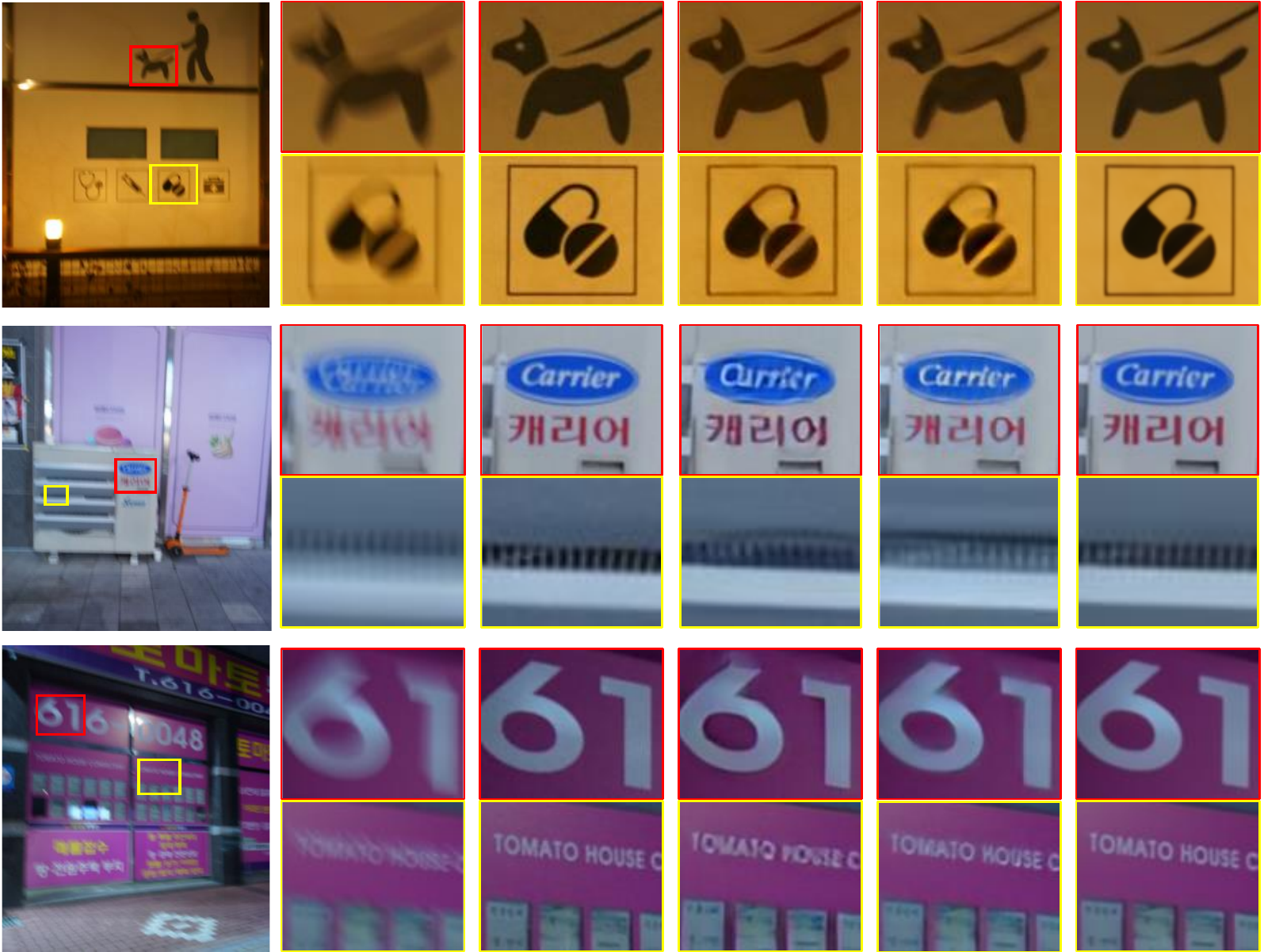}
	\caption{Several examples on the RealBlur test dataset. For clarity, the magnified parts of the resultant images are displayed. From left to right: Blurry images, ground-truth images, and the resultant images obtained by DeblurGAN-v2, SRN, and MIMO-UNet++, respectively.}
	\vspace{-.1cm}
	\label{fig: resultimage_real}
\end{figure*}

\subsection{Loss function}
Likewise with other multi-scale deblurring networks, we use the multi-scale content loss function~\cite{nah2017}, where we found that L1 loss produces better results than MSE loss for our network. The content loss $L_{cont}$ is defined as follows:
\begin{equation}
	L_{cont} = \sum_{k=1}^{K} {\dfrac{1}{t_k } \parallel {\hat{S}}_{k} - S_{k}\parallel_{1}},
\end{equation}
where $K$ is the number of levels. We divide the loss by the number of total elements $t_{k}$ for normalization.

Recent studies also suggest the auxiliary loss terms in addition to the content loss for the performance improvement~\cite{jiao2018look, ignatov2017dslr}. In image enhancement and restoration tasks, auxiliary loss terms that minimize the distance between the input and output in the feature space have been widely used and showed promising results~\cite{zhao2016loss, johnson2016perceptual, ichimura2018spatial, jiang2020focal, Zheng_2020_CVPR}.
Since the purpose of deblurring is to restore the lost high-frequency component, it is essential to reduce the difference in the frequency space.
To this end, we present multi-scale frequency reconstruction~(MSFR) loss function. 
The MSFR loss measures the L1 distance between multi-scale ground-truth and deblurred images in the frequency domain as follows:

\begin{equation}
	L_{MSFR} = \sum_{k=1}^{K} {\dfrac{1}{t_k } \parallel \mathcal{F}({\hat{S}}_{k}) - \mathcal{F}(S_{k})\parallel_{1}},
\end{equation}
where $\mathcal{F}$ denotes the fast Fourier transform (FFT) that transfers image signal to the frequency domain. The final loss function for training our network is determined as follows:
\begin{equation}
	L_{total} = L_{cont} + \lambda L_{MSFR},
\end{equation}
where we experimentally set $\lambda = 0.1$.

\section{Experiments}

\subsection{Dataset and implementation details}
We used the GoPro~\cite{nah2017} and RealBlur~\cite{rim2020real} training datasets for training our models which consist of 2,103 and 3,758 pairs of blurred and sharp images. The GoPro and Real blur test datasets were used for testing, where the number of image pairs are 1,111 and 980, respectively.
For testing on the GoPro test dataset, we trained our model using only the GoPro training dataset.

For every training iteration, we randomly sampled four images and then randomly cropped the sampled images with the size of $ 256 \times 256$. For data augmentation, each patch was horizontally flipped with a probability of 0.5. For deblurring of images in the GoPro dataset, we trained our network for 3,000 epochs which were sufficient for convergence. The learning rate was initially set to $ 10^{-4}$ and decreased by the factor of 0.5 at every 500 epochs. For deblurring of images in the RealBlur dataset, we trained our network for 1,000 epochs, and used the same initial learning rate but decreased it by the factor of 0.5 at every 200 epochs.
Our experiments were conducted on Intel i5-8400 and NVIDIA Titan XP.

\subsection{Performance comparison}

We compared MIMO-UNet with state-of-the-art deblurring networks ~\cite{nah2017, tao2018scale, gao2019dynamic, zhang2019deep, suin2020spatially, park2020multi, purohit2020region, Yuan_2020_CVPR, shen2019human}.
Considering the trade-off between the computational complexity and deblurring accuracy, we evaluated the following three variants of MIMO-UNet: 1) MIMO-UNet employing 8 residual blocks for each EB and DB, 2) MIMO-UNet+ employing 20 residual blocks for each EB and DB, and 3) MIMO-UNet++ estimating the resultant image using MIMO-UNet+ with geometric self-ensemble~\cite{lim2017enhanced}.
The quantitative results on the GoPro test dataset are reported in Table~\ref{table: gopro}.
For a fair comparison, the runtime of the models is provided as the runtime measured using the released test code of each model on our PC (left) and the runtime reported in each paper (right).

\begin{table}[]
	\setlength{\tabcolsep}{4pt} 
	\centering
	
	\begin{tabular}{l|c|c|c|c|c}
		\hline
		Model         & PSNR & SSIM & \multicolumn{2}{c|}{Runtime} & Params.  \\ \hline
		DeepDeblur~\cite{nah2017}      & 29.23    & 0.916  & N/A     & 4.33 & 11.7   \\ 
		SRN~\cite{tao2018scale}   & 30.26       & 0.934    & 0.342& 1.87 & 6.8 \\ 
		PSS-NSC~\cite{gao2019dynamic}      & 30.92     & 0.942      & 0.985 & 1.6 & \underline{2.84} \\ 
		DMPHN~\cite{zhang2019deep}  & 31.20     & 0.945      & 1.061 & 0.424 & 21.7    \\ 
		SAPHN$\dagger$~\cite{suin2020spatially}   & 31.85 & 0.948  & N/A & 0.34 & N/A         \\ 
		SAPHN$\ddagger$~\cite{suin2020spatially}     & 32.02 & 0.953  & N/A & 0.77 & N/A   \\ 
		MT-RNN~\cite{park2020multi} & 31.15 & 0.945  & 0.063 & 0.07 & \textbf{2.6} \\ 
		RADN~\cite{purohit2020region}     & 31.76 & 0.953 & N/A  & 0.038 & N/A \\ 
		SVDN~\cite{Yuan_2020_CVPR} &  29.81 &  0.937   & N/A &   \underline{0.01} & N/A  \\  
		MPRNet~\cite{Zamir_2021_CVPR} &  \underline{32.66} &  \underline{\textbf{0.959}}   & 0.162 &   0.18 & 20.1  \\ \hline 
		MIMO-UNet & 31.73   & 0.951       & \multicolumn{2}{c|}{\textbf{0.008}} & 6.8 \\ 
		MIMO-UNet+ & 32.45  & 0.957   & \multicolumn{2}{c|}{0.017} & 16.1 \\
		MIMO-UNet++ & \textbf{32.68}  & \underline{\textbf{0.959}}   & \multicolumn{2}{c|}{0.040} & 16.1 \\\hline
		
	\end{tabular}
	\caption{The average PSNR and SSIM on the GoPro test dataset. The SAPHNs with $\dagger$ and $\ddagger$ denote the models with and without offsets, respectively. We employ stacked(4) version for DMPHN. The runtime and parameters are expressed in seconds and millions.} 
	\vspace{-.1cm}
	\label{table: gopro}
\end{table}

MIMO-UNet+ and MIMO-UNet++ were slower than MIMO-UNet but still performed deblurring in 0.014s and 0.040s, respectively. The average PSNR of MIMO-UNet++ was obtained as 32.68 dB. MIMO-UNet showed the average processing time of 0.008s and the average PSNR of 31.73 dB. These three models demonstrate the best trade-off between the accuracy and computational complexity as shown in Figure~\ref{fig: graph}.\footnote{Runtime measured using GPU synchronization mode can be find in on our website.}
Due to the stacked sub-networks, DeepDeblur, SRN, PSS-NSC, DMPHN, and SAPHN required large computational costs as shown in Table~\ref{table: gopro}. 
Compared with these methods, MIMO-UNet+ was faster but achieved still higher PSNR scores.
Although SRN, PSS-NSC, and MT-RNN employ fewer parameters than the proposed methods, these methods repetitively use parameters in the procedure, and therefore they are slower than the our slowest model MIMO-UNet++.
Especially, MIMO-UNet++ was 4.05 times faster and 0.02 dB higher in terms of PSNR compared to MPRNet that is the best method among the conventional methods.
The single network-based methods, such as RADN and SVDN, achieved high runtime efficiency compared to the stacked sub-networks.
However, MIMO-UNet outperforms SVDM, and MIMO-UNet+ outperforms RADN, in terms of both runtime and PSNR.
To validate the effectiveness of the proposed method on the real case scenario, we also evaluated our methods on the recent RealBlur dataset~\cite{rim2020real}.
As listed in Table~\ref{table: realblur}, MIMO-UNet++ recorded the best and the second best performance in terms of PSNR and SSIM, respectively.
The several resultant images from the GoPro and RealBlur test datasets are shown in Figure~\ref{fig: resultimage} and Figure~\ref{fig: resultimage_real}, respectively.
For the reproduction of results, we used the author-released network models trained on each dataset, \textit{i.e.}, SRN, PSS-NSC, DMPHN, MT-RNN, and MPRNet were used for the GoPro dataset, and DeblurGAN-v2 and SRN for the RealBlur dataset, respectively.
Although the resultant images obtained by the conventional networks exhibit much less blur compared to the input blurry images, local details and structures were not sufficiently deblurred as can be noticed from the magnified image regions, whereas our method produced sharper images.

\subsection{Ablation study} \label{subsec: abl}

\begin{table}[]
	\begin{center}
		\setlength{\tabcolsep}{4pt} 
		\begin{tabular}{l|cc}
			\hline
			Model & PSNR & SSIM \\  \hline
			DeblurGAN-v2~\cite{kupyn2019deblurgan} & 29.69 & 0.870 \\
			SRN~\cite{tao2018scale} & 31.38 & 0.909 \\ 
			MPRNet~\cite{Zamir_2021_CVPR} & 31.76 & \textbf{0.922} \\ \hline
			MIMO-UNet+ & \underline{31.92} & 0.919 \\
			MIMO-UNet++ & \textbf{32.05} & \underline{0.921} \\ \hline
			
		\end{tabular}
	\end{center}
	\caption{The average PSNR and SSIM on the RealBlur test dataset~\cite{rim2020real}.}
	\label{table: realblur}
\end{table}

\begin{table}[]
	\begin{center}
		\setlength{\tabcolsep}{4pt} 
		\begin{tabular}{l|ccc}
			\hline
			Method & Concat. & Element-wise sum & FAM \\ \hline
			PSNR   & 31.66 & 31.60  & \textbf{31.73} \\ \hline    
		\end{tabular}
	\end{center}
	\caption{Ablation studies on FAM.}
	\label{table: fam}
\end{table}

\begin{table}[]
	\begin{center}
		\begin{tabular}{cccc|c|c}
			\hline
			MISE        & MOSD        & AFF        & MSFR        & PSNR & Params. \\\hline
			&            &            &            & 31.16 & 6.46\\
			\checkmark &            &            &            & 31.17 & 6.72\\
			& \checkmark &            &            & 31.33  & 6.47\\
			&            & \checkmark &            & 31.33  & 6.54\\
			\checkmark & \checkmark &            &            & 31.38  & 6.73\\
			& \checkmark & \checkmark  &            & 31.38  & 6.54 \\
			\checkmark &            & \checkmark &            & 31.39  & 6.80 \\
			\checkmark & \checkmark & \checkmark &            & 31.46  & 6.81 \\
			\checkmark & \checkmark & \checkmark & \checkmark & \textbf{31.73} & 6.81\\\hline
		\end{tabular}
	\end{center}
	
	\caption{Effectiveness of different components of MIMO-UNet on the GoPro test dataset.}
	\vspace{-.1cm}
	\label{table: ablation}
\end{table}

We conducted experiments to analyze the effectiveness of each component of MIMO-UNet on the GoPro test dataset. 
First, we evaluated the effectiveness of different feature fusion methods in MISE.
The proposed FAM was compared with the conventional fusion methods: concatenation and element-wise sum, and achieved the highest performance as listed in Table~\ref{table: fam}.
Second, we tested MIMO-UNet without MOSD, MISE, AFF, and/or MSFR. 
For comparison, a baseline model was trained without using any of the four components, resulting the average PSNR of 31.16 dB.
As shown in Table~\ref{table: ablation}, compared with the baseline model, MOSD improved PSNR by 0.17 dB. The standalone use of MISE showed a marginal effect because multi-scale information is difficult to be used in a simple U-Net. However, when used with MOSD, MISE contributed to the further performance improvement of PSNR by 0.05 dB. AFF improved PSNR by 0.17 dB compared to the baseline model, and the performance gain was further increased to 0.23 dB when AFF was used with MISE. With MISE, MOSD, and AFF, the network achieved 0.30 dB higher PSNR, and finally, the network trained using MSFR achieved 0.57 dB higher PSNR compared to the baseline.

\begin{figure}[!t]
	\centering
	\includegraphics[width=.9\columnwidth]{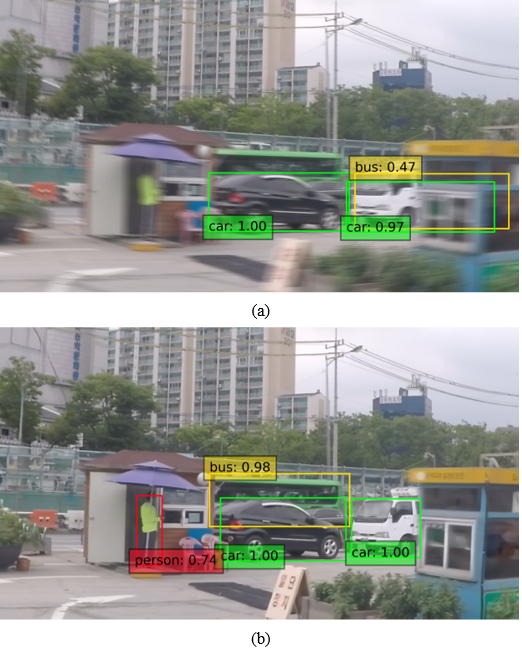}
	\caption{The examples of object detection result from (a) blurry image and (b) resultant image obtained by MIMO-UNet++.}
	\label{fig: detection}
\end{figure}

\subsection{Object detection performance evaluation}

Single image deblurring can also boost the performance of computer vision tasks when used as a preprocessing technique. Object detection is one of the best examples in which single image deblurring can be used to improve the performance. With the advances in the CNNs, object detection methods have adopted CNNs and achieved significant improvements~\cite{lin2017focal, Redmon_2016_CVPR}. However, most of these methods assume blur-free input images, and therefore they often fail to detect objects in blurry images. Figure~\ref{fig: detection}(a) illustrates the failure case of PFPNet~\cite{kim2018parallel}, which is one of the state-of-the-art object detectors, in detecting objects from a blurry image, depicting its vulnerability to blurry inputs. When the same PFPNet was applied to the deblurred image obtained using MIMO-UNet++, many of the false negative examples could be successfully detected as shown in Figure~\ref{fig: detection}(b).

\begin{figure}[!t]
	\centering
	\includegraphics[width=0.9\columnwidth]{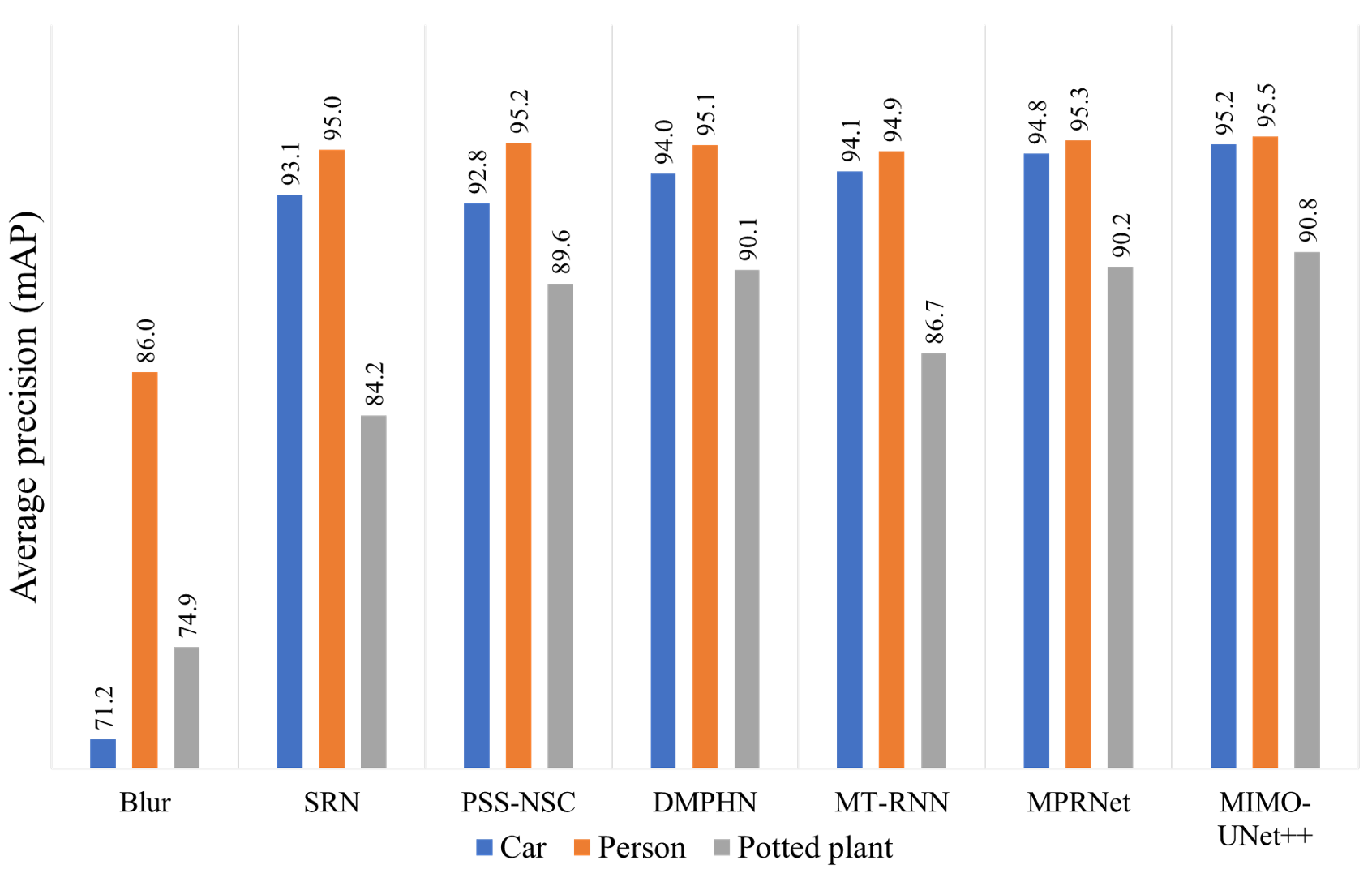}
	\caption{Object detection performance evaluation. Following the measurement~\cite{Kupyn_2018_CVPR}, we use the bounding boxes obtained from the sharp images as the ground-truth.}
	\label{fig: detection_graph}
\end{figure}

Last, we compared the proposed MIMO-UNet++ with the other deblurring techniques in terms of their effectiveness in the object detection task as preprocessing. Similar to the previous experiment, PSS-NSC and DMPHN with the author-provided codes were used for comparison. 
Although PFPNet was trained using the PASCAL VOC dataset~\cite{everingham2010pascal} that contains 20 different classes, the blurry images in the GoPro dataset primarily contain only three classes among them, \textit{i.e.}, car, person, and potted plant. Therefore, the average precision~(AP) of each object class was measured for the performance evaluation. As shown in Figure~\ref{fig: detection_graph}, the proposed MIMO-UNet++ resulted in the best performance in object detection. Moreover, since the proposed method recorded the fastest execution time, it is most suitable as a preprocessing technique for object detection.

\section{Conclusion}
In this paper, we proposed a fast and accurate image deblurring network. Instead of stacking multiple sub-networks for coarse-to-fine deblurring, we presented a single U-Net that has distinct features, enabling much simpler but more effective coarse-to-fine deblurring. The encoder of the network is modified to take multi-scale input images and combine features from different sources. The decoder of the network is also changed to output multi-scale deblurred images during decoding such that coarse-to-fine deblurring can be better performed. A feature fusion method is also introduced to asymmetrically combine multi-scale features for dynamic image deblurring. The experimental results demonstrate that our method outperforms the other conventional methods in regard to the speed and accuracy trade-off.

\section*{Acknowlegement}
This work was supported by Samsung Electronics Co., Ltd (IO201210-08026-01)

{\small
\bibliographystyle{ieee_fullname}
\bibliography{egbib}

\begin{thebibliography}{10}\itemsep=-1pt

\bibitem{chakrabarti2016neural}
Ayan Chakrabarti.
\newblock A neural approach to blind motion deblurring.
\newblock In {\em Eur. Conf. Comput. Vis.}, pages 221--235, 2016.

\bibitem{everingham2010pascal}
Mark Everingham, Luc Van~Gool, Christopher~KI Williams, John Winn, and Andrew
  Zisserman.
\newblock The pascal visual object classes (voc) challenge.
\newblock {\em Int. J. Comput. Vis.}, 88(2):303--338, 2010.

\bibitem{fergus2006removing}
Rob Fergus, Barun Singh, Aaron Hertzmann, Sam~T Roweis, and William~T Freeman.
\newblock Removing camera shake from a single photograph.
\newblock In {\em ACM SIGGRAPH 2006 Papers}, pages 787--794. 2006.

\bibitem{franke2000real}
Uwe Franke and Armin Joos.
\newblock Real-time stereo vision for urban traffic scene understanding.
\newblock In {\em Proceedings of the IEEE Intelligent Vehicles Symposium 2000
  (Cat. No. 00TH8511)}, pages 273--278. IEEE, 2000.

\bibitem{gao2019dynamic}
Hongyun Gao, Xin Tao, Xiaoyong Shen, and Jiaya Jia.
\newblock Dynamic scene deblurring with parameter selective sharing and nested
  skip connections.
\newblock In {\em IEEE Conf. Comput. Vis. Pattern Recog.}, pages 3848--3856,
  2019.

\bibitem{guo2018hierarchical}
Chunle Guo, Chongyi Li, Jichang Guo, Runmin Cong, Huazhu Fu, and Ping Han.
\newblock Hierarchical features driven residual learning for depth map
  super-resolution.
\newblock {\em IEEE Trans. Image Process.}, 28(5):2545--2557, 2018.

\bibitem{hradivs2015convolutional}
Michal Hradi{\v{s}}, Jan Kotera, Pavel Zemc{\i}k, and Filip {\v{S}}roubek.
\newblock Convolutional neural networks for direct text deblurring.
\newblock In {\em Brit. Mach. Vis. Conf.}, volume~10, page~2, 2015.

\bibitem{ichimura2018spatial}
Naoyuki Ichimura.
\newblock Spatial frequency loss for learning convolutional autoencoders.
\newblock {\em arXiv preprint arXiv:1806.02336}, 2018.

\bibitem{ignatov2017dslr}
Andrey Ignatov, Nikolay Kobyshev, Radu Timofte, Kenneth Vanhoey, and Luc
  Van~Gool.
\newblock {DSLR}-quality photos on mobile devices with deep convolutional
  networks.
\newblock In {\em Int. Conf. Comput. Vis.}, pages 3277--3285, 2017.

\bibitem{jiang2020focal}
Liming Jiang et~al.
\newblock Focal frequency loss for generative models.
\newblock {\em arXiv preprint arXiv:2012.12821}, 2020.

\bibitem{jiao2018look}
Jianbo Jiao, Ying Cao, Yibing Song, and Rynson Lau.
\newblock Look deeper into depth: Monocular depth estimation with semantic
  booster and attention-driven loss.
\newblock In {\em Eur. Conf. Comput. Vis.}, pages 53--69, 2018.

\bibitem{johnson2016perceptual}
Justin Johnson, Alexandre Alahi, and Li Fei-Fei.
\newblock Perceptual losses for real-time style transfer and super-resolution.
\newblock In {\em Eur. Conf. Comput. Vis.}, pages 694--711, 2016.

\bibitem{kim2018parallel}
Seung-Wook Kim, Hyong-Keun Kook, Jee-Young Sun, Mun-Cheon Kang, and Sung-Jea
  Ko.
\newblock Parallel feature pyramid network for object detection.
\newblock In {\em Eur. Conf. Comput. Vis.}, pages 234--250, 2018.

\bibitem{Kupyn_2018_CVPR}
Orest Kupyn, Volodymyr Budzan, Mykola Mykhailych, Dmytro Mishkin, and Jiří
  Matas.
\newblock Deblurgan: Blind motion deblurring using conditional adversarial
  networks.
\newblock In {\em IEEE Conf. Comput. Vis. Pattern Recog.}, pages 8183--8192,
  2018.

\bibitem{kupyn2019deblurgan}
Orest Kupyn, Tetiana Martyniuk, Junru Wu, and Zhangyang Wang.
\newblock Deblurgan-v2: Deblurring (orders-of-magnitude) faster and better.
\newblock In {\em Int. Conf. Comput. Vis.}, pages 8878--8887, 2019.

\bibitem{lim2017enhanced}
Bee Lim, Sanghyun Son, Heewon Kim, Seungjun Nah, and Kyoung Mu~Lee.
\newblock Enhanced deep residual networks for single image super-resolution.
\newblock In {\em IEEE Conf. Comput. Vis. Pattern Recog. Worksh.}, pages
  136--144, 2017.

\bibitem{lin2017focal}
Tsung-Yi Lin, Priya Goyal, Ross Girshick, Kaiming He, and Piotr Doll{\'a}r.
\newblock Focal loss for dense object detection.
\newblock In {\em IEEE Conf. Comput. Vis. Pattern Recog.}, pages 2980--2988,
  2017.

\bibitem{Liu2016CSVT}
S. {Liu}, H. {Wang}, J. {Wang}, and C. {Pan}.
\newblock Blur-kernel bound estimation from pyramid statistics.
\newblock {\em IEEE Trans. Circuit Syst. Video Technol.}, 26(5):1012--1016,
  2016.

\bibitem{michaeli2014blind}
Tomer Michaeli and Michal Irani.
\newblock Blind deblurring using internal patch recurrence.
\newblock In {\em Eur. Conf. Comput. Vis.}, pages 783--798, 2014.

\bibitem{nah2017}
Seungjun Nah, Tae~Hyun Kim, and Kyoung~Mu Lee.
\newblock Deep multi-scale convolutional neural network for dynamic scene
  deblurring.
\newblock In {\em IEEE Conf. Comput. Vis. Pattern Recog.}, pages 3883--3891,
  2017.

\bibitem{pang2019efficient}
Yanwei Pang, Tiancai Wang, Rao~Muhammad Anwer, Fahad~Shahbaz Khan, and Ling
  Shao.
\newblock Efficient featurized image pyramid network for single shot detector.
\newblock In {\em IEEE Conf. Comput. Vis. Pattern Recog.}, pages 7336--7344,
  2019.

\bibitem{park2020multi}
Dongwon Park, Dong~Un Kang, Jisoo Kim, and Se~Young Chun.
\newblock Multi-temporal recurrent neural networks for progressive non-uniform
  single image deblurring with incremental temporal training.
\newblock In {\em Eur. Conf. Comput. Vis.}, pages 327--343, 2020.

\bibitem{purohit2020region}
Kuldeep Purohit and AN Rajagopalan.
\newblock Region-adaptive dense network for efficient motion deblurring.
\newblock In {\em AAAI}, pages 11882--11889, 2020.

\bibitem{Redmon_2016_CVPR}
Joseph Redmon, Santosh Divvala, Ross Girshick, and Ali Farhadi.
\newblock You only look once: Unified, real-time object detection.
\newblock In {\em IEEE Conf. Comput. Vis. Pattern Recog.}, pages 779--788, June
  2016.

\bibitem{rim2020real}
Jaesung Rim, Haeyun Lee, Jucheol Won, and Sunghyun Cho.
\newblock Real-world blur dataset for learning and benchmarking deblurring
  algorithms.
\newblock In {\em Eur. Conf. Comput. Vis.}, pages 184--201, 2020.

\bibitem{ronneberger2015u}
Olaf Ronneberger, Philipp Fischer, and Thomas Brox.
\newblock U-net: Convolutional networks for biomedical image segmentation.
\newblock In {\em International Conference on Medical Image Computing and
  Computer-assisted Intervention}, pages 234--241, 2015.

\bibitem{schuler2015learning}
Christian~J Schuler, Michael Hirsch, Stefan Harmeling, and Bernhard
  Sch{\"o}lkopf.
\newblock Learning to deblur.
\newblock {\em IEEE Trans. Pattern Anal. Mach. Intell.}, 38(7):1439--1451,
  2015.

\bibitem{shen2019human}
Ziyi Shen, Wenguan Wang, Xiankai Lu, Jianbing Shen, Haibin Ling, Tingfa Xu, and
  Ling Shao.
\newblock Human-aware motion deblurring.
\newblock In {\em Int. Conf. Comput. Vis.}, pages 5572--5581, 2019.

\bibitem{suin2020spatially}
Maitreya Suin, Kuldeep Purohit, and AN Rajagopalan.
\newblock Spatially-attentive patch-hierarchical network for adaptive motion
  deblurring.
\newblock In {\em IEEE Conf. Comput. Vis. Pattern Recog.}, pages 3606--3615,
  2020.

\bibitem{sun2015learning}
Jian Sun, Wenfei Cao, Zongben Xu, and Jean Ponce.
\newblock Learning a convolutional neural network for non-uniform motion blur
  removal.
\newblock In {\em IEEE Conf. Comput. Vis. Pattern Recog.}, pages 769--777,
  2015.

\bibitem{tao2018scale}
Xin Tao, Hongyun Gao, Xiaoyong Shen, Jue Wang, and Jiaya Jia.
\newblock Scale-recurrent network for deep image deblurring.
\newblock In {\em IEEE Conf. Comput. Vis. Pattern Recog.}, pages 8174--8182,
  2018.

\bibitem{thorpe2013coprime}
Christopher Thorpe, Feng Li, Zijia Li, Zhan Yu, David Saunders, and Jingyi Yu.
\newblock A coprime blur scheme for data security in video surveillance.
\newblock {\em IEEE Trans. Pattern Anal. Mach. Intell.}, 35(12):3066--3072,
  2013.

\bibitem{Yuan_2020_CVPR}
Yuan Yuan, Wei Su, and Dandan Ma.
\newblock Efficient dynamic scene deblurring using spatially variant
  deconvolution network with optical flow guided training.
\newblock In {\em IEEE Conf. Comput. Vis. Pattern Recog.}, pages 3555--3564,
  2020.

\bibitem{Zamir_2021_CVPR}
Syed~Waqas Zamir, Aditya Arora, Salman Khan, Munawar Hayat, Fahad~Shahbaz Khan,
  Ming-Hsuan Yang, and Ling Shao.
\newblock Multi-stage progressive image restoration.
\newblock In {\em IEEE Conf. Comput. Vis. Pattern Recog.}, pages 14821--14831,
  2021.

\bibitem{zhang2019deep}
Hongguang Zhang, Yuchao Dai, Hongdong Li, and Piotr Koniusz.
\newblock Deep stacked hierarchical multi-patch network for image deblurring.
\newblock In {\em IEEE Conf. Comput. Vis. Pattern Recog.}, pages 5978--5986,
  2019.

\bibitem{zhao2016loss}
Hang Zhao, Orazio Gallo, Iuri Frosio, and Jan Kautz.
\newblock Loss functions for image restoration with neural networks.
\newblock {\em IEEE Trans. Comput. Imag.}, 3(1):47--57, 2016.

\bibitem{Zheng_2020_CVPR}
Bolun Zheng, Shanxin Yuan, Gregory Slabaugh, and Ales Leonardis.
\newblock Image demoireing with learnable bandpass filters.
\newblock In {\em IEEE Conf. Comput. Vis. Pattern Recog.}, pages 3636--3645,
  2020.

\end{thebibliography}
}

\end{document}